\documentclass[11pt,a4paper]{article}
\usepackage{authblk} % this must be a second line in the file
\usepackage{acl2019}

\usepackage{times}
\usepackage{latexsym}
\usepackage{url}
\usepackage{booktabs}
\usepackage{multicol}
\usepackage{multirow}
\usepackage{paralist}
\usepackage[utf8]{inputenc}
\usepackage{mathtools}
\usepackage{colortbl}
\usepackage{amssymb}
\usepackage{todonotes}
\usepackage{graphicx}
\usepackage{csquotes}
\usepackage{array}
\usepackage{caption}
\usepackage{enumitem}
\usepackage{amsmath}
\usepackage{booktabs}

\usepackage{breakurl}

\DeclareMathOperator{\arccosh}{arcosh}

\usepackage{ctable}
\newcolumntype{P}[1]{>{\centering\arraybackslash}p{#1}}
\newcolumntype{M}[1]{>{\centering\arraybackslash}m{#1}}

\aclfinalcopy % Uncomment this line for the final submission
\def\aclpaperid{1381} %  Enter the acl Paper ID here

\title{On the Compositionality Prediction of Noun Phrases\\ using Poincar{\'e} Embeddings}

\author[$\dag$]{\textbf{Abhik Jana}}
\author[$\ddag$]{\textbf{Dmitry Puzyrev}}
\author[$\star,\S$]{\textbf{Alexander Panchenko}}
\author[$\dag$]{\textbf{Pawan Goyal}}
\author[$\S$]{\\\textbf{Chris Biemann}}
\author[$\dag$]{\textbf{Animesh Mukherjee}}

\affil[$\dag$]{Indian Institute of Technology Kharagpur, Kharagpur, India}
\affil[$\ddag$]{National Research University Higher School of Economics, Moscow, Russia}
\affil[$\star$]{Skolkovo Institute of Science and Technology, Moscow, Russia}
\affil[$\S$]{Universit{\"a}t Hamburg, Hamburg, Germany}

\affil[ ]{\href{mailto:abhik.jana@iitkgp.ac.in}{abhik.jana@iitkgp.ac.in, \{pawang,animeshm\}@cse.iitkgp.ac.in}}
\affil[ ]{\href{mailto:dapuzyrev@edu.hse.ru}{dapuzyrev@edu.hse.ru}}

\affil[ ]{\href{mailto:panchenko@informatik.uni-hamburg.de}{\{panchenko,biemann\}@informatik.uni-hamburg.de}}

\date{}

\begin{document}
\maketitle
\begin{abstract}
The compositionality degree of multiword expressions indicates to what extent the meaning of a phrase can be derived from the meaning of its constituents and their grammatical relations. Prediction of (non)-compositionality is a task that has been frequently addressed with distributional semantic models. We introduce a novel technique to blend hierarchical information with distributional information for predicting compositionality. In particular, we use hypernymy information of the multiword and its constituents encoded in the form of the recently introduced Poincar{\'e} embeddings in addition to the distributional information to detect compositionality for noun phrases. Using a weighted average of the distributional similarity and a Poincar{\'e} similarity function, we obtain consistent and substantial, statistically significant improvement across three gold standard datasets over state-of-the-art models based on distributional information only. Unlike traditional approaches that solely use an unsupervised setting, we have also framed the problem as a supervised task, obtaining comparable improvements. Further, we publicly release our Poincar{\'e} embeddings, which are trained on the output of handcrafted lexical-syntactic patterns on a large corpus.      
\end{abstract}

\section{Introduction }
 An important challenge in Natural Language Processing is to represent words, phrases, and larger spans in a way that reflects their meaning.  Compositionality is one of the strongest assumptions in semantics, stating that the meaning of larger units can be derived from their smaller parts and their contextual relation. However, for idiomatic phrases, this assumption does not hold true as the meaning of the whole phrase may not be related to their parts in a straightforward fashion. The meaning of the phrases like `data format', `head teacher', `green tree' can easily be understood from the constituent words whereas the semantics of the idiomatic phrases like `couch potato', `rat race', `nut case' are non-compositional, i.e., refer to a different meaning than their parts suggest. 
 
 In this work, we address compositionality prediction, which is the task of assigning a numerical score to a phrase indicating the extent to which the meaning of the phrase can be derived from the meanings of its constituent words. To motivate its importance, e.g., in machine translation, non-compositional phrases must be translated as a unit; in word sense disambiguation,  assigning one of the constituent word's senses to the whole phrase should be avoided for idiomatic phrases; semantic parsing also requires to correctly identify complex predicates and their arguments in this way.  
 
 A significant amount of effort has gone into operationalizing dense-vector distributional semantic models (DSMs) of different flavors such as  count-based models~(\newcite{baldwin2003empirical,venkatapathy2005measuring,mccarthy2007detecting}), word embeddings based on word2vec (both CBOW and SkipGram) and similar~(\newcite{reddy2011empirical,salehi2014using,P16-1187,cordeiro2018unsupervised}), and multi-sense skip-gram models for compositionality prediction~\cite{salehi2015word}.  All these attempts are based on the hypothesis that the composition of the representation of constituent words will be closer to the representation of the entire phrase in case of compositional phrases as compared to the non-compositional ones~\cite{choueka1988looking}. 
 
 Observing that the distributional information alone is not enough for precise compositionality prediction, we propose to utilize hypernymy information, hypothesizing that, for compositional phrases, the hypernym of the whole phrase is semantically closer to the hypernyms of one of the constituent words (head words) as compared to the non-compositional phrases. For example, `art school' and `school' have one common hypernym `educational institution' whereas `hot dog' has no common hypernym with `hot' or `dog', apart from very abstract concepts such as `physical entity'. Of course, this only holds for noun phrases, where taxonomic relations between nouns apply.

 To represent hypernymy information we use Poincar{\'e} embeddings \cite{NIPS2017_7213} for learning hierarchical representations of symbolic data by embedding them into a hyperbolic space. To this end,  we extract hyponym-hypernym pairs by applying well-known lexical-syntactic patterns proposed by \citet{Hearst:1992:AAH:992133.992154} on a large corpus and train Poincar{\'e} embeddings on a list of hyponym-hypernym pairs.
 
 Relying on two types of representations, i.e., dense vectors in the Euclidean space and the novel hyperbolic Poincar{\'e} embeddings, we interpolate their similarity predictions in a novel compositionality score metric that takes both distributional and hypernymy information into account. We evaluate our proposed metric on three well-accepted English datasets, i.e.,  Reddy \cite{reddy2011empirical}, Reddy++ \cite{ramisch2016naked} and Farahmand \cite{W15-0904}, demonstrating a performance boost when including hyperbolic embeddings by 2-4\% absolute points across all datasets. 
 
 In particular, our work contains the three following 
 \textbf{contributions}:
 
\begin{enumerate}
     \item  We devise a straightforward and efficient approach for combining distributional and hypernymy information for the task of noun phrase compositionality prediction. As far as we are aware, this is the first application of Poincar{\'e} embeddings to this task.
     \item We demonstrate consistent and significant improvements on benchmark datasets in unsupervised and supervised settings.
     \item We publicly release our Poincar{\'e} embeddings trained on pattern extractions on a very large corpus.

 \end{enumerate}
 
\section{Related Work} 
\label{sec:related}

Some of the initial efforts on compositionality prediction were undertaken by~\citet{baldwin2003empirical}, who use LSA to calculate the similarity between a phrase and its components, whereas~\citet{venkatapathy2005measuring} extend this idea with collocation features (e.g., phrase frequency, point-wise mutual information). Researchers also tried to identify  non-compositionality in verb-noun phrases using syntax~\cite{cook2007pulling} and selectional preferences~\cite{mccarthy2007detecting}.
Attempts to examine the possibility to derive the semantics of a compound or multiword expression from its parts have been researched extensively \citep{mccarthy2003detecting,mitchell2008vector,tratz2010isi}.
~\citet{reddy2011empirical} define a compositionality score and use different vector operations to estimate the semantic distance between a phrase and its individual components. 
Some of the investigations are made for compositionality detection using representation learning of word embeddings~\citep{socher2012semantic,salehi2015word}.
~\citet{salehi2014using} also show that distributional similarity over multiple languages can help in improving the quality of compositionality prediction.

In a recent attempt,~\citet{yazdani2015learning} tries to learn semantic composition and finds that complex functions such as polynomial projection and neural networks can model semantic composition more effectively than the commonly used additive and multiplicative functions. 
\citet{kiela2013detecting} detect non-compositionality using concepts of mutual information.~\citet{lioma2015non} replace the context vectors with language models and compute their Kullback–Leibler divergence to approximate their semantic distance. 
In another stream, researchers have also attempted to classify idiomatic vs. non-idiomatic expressions in different languages considering the context of the expressions~\cite{flor2018catching,bizzoni2018finding,peng2018distributional}, see also a respective shared task \citep{biemann-giesbrecht-2011-distributional}.        
In one of the recent attempts,~\citet{P16-1187} conduct an analysis of several DSMs (word2vec, GloVe, PPMI) with variations of hyper-parameters and produce the state-of-the-art results in the compositionality prediction task, which is extended further for different languages by~\citet{cordeiro2018unsupervised}. We take their work as our baseline and carry forward our investigation to improve the state-of-the-art performance by introducing the hyponymy-hypernymy information in the form of Poincar\'e embeddings.

\citet{le2019inferring} and \citet{aly2019} also showed usefulness the use of  Poincar\'e embeddings: in their case for inducing taxonomies from the text. In both works, hyperbolic embeddings are trained using relations harvested using Hearst patterns, like in our work. The usefulness of hyperbolic embeddings was also shown beyond text processing: \citet{khrulkov2019hyperbolic} successfully applied them for hierarchical relations in image classification tasks.

\section{Methodology}
\label{model}
 Our aim is to produce a compositionality score for a given two-word noun phrase $w_1 w_2$. As per our hypothesis, the proposed compositionality score metric has two components: one component takes care of the extent of the distributional similarity between the phrase and the composition of constituent words. The second component captures hypernymy-based similarity obtained through Poincar{\'e} embeddings~\cite{NIPS2017_7213}. The rationale behind this is that replacing a word with its hypernym should yield phrases with similar meaning for compositional cases, dissimilar phrases otherwise (e.g., a `red herring' is not similar to `red fish'). 
 
\paragraph{Distributional component:} For the first component, we follow the scheme prescribed by \citet{P16-1187}, relying on the state-of-the-art DSM model and the score metric ($Score_{D}$) proposed in that work. The metric $Score_{D}$ is defined as,
\begin{equation}\label{eq1}
Score_{D} (w_1w_2) = \cos ( v(w_1w_2), v(w_1+w_2)),
\end{equation}
where 
\begin{equation}
v(w_1+w_2) = \frac{v(w_1)}{\|v(w_1)\|} +  \frac{v(w_2)}{\|v(w_2)\|},
\end{equation}
and $v(w)$ is the vector representation of $w$ obtained from the DSM, $||.||$ is the L2-norm. For the composition of two component word vectors, we use the additive model, which is well-accepted in the literature~\cite{mitchell2010composition}. 
 
\paragraph{Hypernymy component:} For the second component, we prepare Poincar{\'e} embeddings. The Poincar{\'e} embedding as introduced by \citet{NIPS2017_7213} is a very recent approach to learn hierarchical representations of symbolic data by embedding them into the hyperbolic space. The underlying hyperbolic geometry helps to learn parsimonious representations of symbolic data by simultaneously capturing hierarchy and similarity. As per this proposed Poincar{\'e} ball model, let 
\begin{equation}
\beta^d= \{x \in \mathbb{R}\quad : \quad \Arrowvert x \Arrowvert < 1\}
\end{equation}
be the open $d$-dimensional unit ball, where $\| . \|$ denotes the Euclidean norm. 
 
 The list of hyponym-hypernym pairs was obtained by applying lexical-syntactic patterns described by~\newcite{Hearst:1992:AAH:992133.992154} on the corpus prepared by~\newcite{panchenko2016taxi}. This corpus is a concatenation of the English Wikipedia (2016 dump), Gigaword \cite{parker}, ukWaC \cite{Ferraresi} and English news corpora from the Leipzig Corpora Collection \cite{goldhahn}. The lexical-syntactic patterns proposed by \citet{Hearst:1992:AAH:992133.992154} and further extended and implemented in the form of FSTs by \citet{panchenko2012semantic}\footnote{\url{https://zenodo.org/record/3234817}}
 for extracting (noisy) hyponym-hypernym pairs are given as follows -- (i) \textit{such} NP \textit{as} NP, NP[,] \textit{and/or} NP; (ii) NP \textit{such as} NP, NP[,] \textit{and/or} NP; (iii) NP, NP [,] \textit{or other} NP; (iv) NP, NP [,] \textit{and other} NP; (v) NP, \textit{including} NP, NP [,] \textit{and/or} NP; (vi) NP, \textit{especially} NP, NP [,] \textit{and/or} NP.

 Pattern extraction on the corpus yields a list of 27.6 million hyponym-hypernym pairs along with the frequency of their occurrence in the corpus. We normalize the frequency of each hyponym-hypernym pair by dividing it by the logarithm of the global frequency of the hypernym in the list, which realizes a TF-IDF \cite{sparck1972statistical} weighting, to downrank noisy extractions with frequent pattern-extracted `hypernyms' such as `problem, issue, bit'. 

Further, we sort the list of hyponym-hypernym pairs with respect to their the normalized frequency. As the Poincar{\'e} embedding method takes as input a list of hyponym-hypernym pairs, we first prepare a list by adding top $k$ pairs (based on normalized frequency) where the noun phrases or component words present in the gold-standard dataset exist as hyponym or hypernym. Note that we embed noun phrases as extracted by the patterns as units, i.e. a term like ``educational institution'' will get its own embedding if it appears in the pattern extractions as an NP. This list is quite sparse and therefore the hyperbolic space is not rich enough to produce good results (see Section~\ref{ER}).

In order to circumvent this problem, we further populate the above list by appending the top $m$ percent pairs from the complete sorted list of hyponym-hypernym pairs we prepared earlier. Next, we use this expanded list as input to prepare Poincar{\'e} embeddings. 

\paragraph{Hyperparameters for training Poincar{\'e} model:} For both the unsupervised and the supervised setup we maintain the  following settings for the training of the Poincar{\'e} model unless otherwise stated: vector dimensionality $d$ = 50, number of negative samples = 2, learning rate = 0.1, coefficient used for L2-regularization while training~=~1, and number of epochs to use for burn-in initialization = 10.

\subsection{Unsupervised Setup}
The Poincar{\'e} distance between points $x, y \in \beta^d$ is defined in the following way:

 \begin{equation}
     d(x,y)=\arccosh{ \left( 1+2\frac{||x-y||^2}{(1-\|x\|^2)(1-\|y\|^2)}\right)}.
 \end{equation}

Poincar{\'e} similarity score $Score_P$ is derived from the Poincar{\'e} distance as 
\begin{equation}
     Score_P (x,y) = \frac {1}{1+d(x,y)}. 
\end{equation}

Let $w_1 w_2$ be the noun phrase for which we compute the compositionality score. Further let $H_{w_1 w_2}$ be the set of top $k$ hypernyms of the phrase $w_1 w_2$ and $H_{w_1}$, $H_{w_2}$ be the set of top $k$ hypernyms of the constituent words $w_1$ and $w_2$, respectively. Our proposed compositionality score metric $Score(w_1 w_2)$ is defined as follows:
\begin{equation}
\begin{split}
Score(w_1w_2) = (1- \alpha) Score_{D} (w_1w_2) + \\ \alpha  \max_{  \substack{a \in H_{w_1w_2}\\  b \in H_{w_1\textcolor{white}{w_2}}\\  c \in H_{w_2\textcolor{white}{w_2}}}} (Score_P (v(a), v(b)+v(c)) ),  
\end{split}
\end{equation}
where $v(w)$ indicates the vector representation of the word $w$ and $\alpha$ is used to set the relative weight of the two components.

 \subsection{Supervised Setup}
 \label{sp}
 We explore the utility of hierarchical information encoded in Poincar\'e embeddings for the task of compositionality prediction in a supervised setup as well. As our aim is to predict a compositionality score, we employ several regression techniques like Support Vector Regression~\cite{Drucker96supportvector}, Kernel Ridge Regression~\cite{vovk2013kernel}, $k$-Nearest Neighbours Regression~\cite{altman1992introduction}, Partial Least Squares Regression (PLS)~\cite{abdi2007partial} etc. We randomly split the full dataset into a 75\% training set and a 25\% test set, and experiment on 25 such random splits. For each split, we plugin the concatenation of the vector representation of the noun phrase as well as the component words. The supervised predicted score is
\begin{equation} 
 \label{se}
\begin{split}
Score_{S}(w_1w_2)  = (1-\alpha) \cdot & Score_{DS}(w_1w_2) + \\
  \alpha \cdot & Score_{PS}(w_1w_2),
\end{split}
\end{equation}
where $Score_{DS}(w_1w_2)$ is the predicted score when we plugin the vectors from DSMs into the regression model and $Score_{PS}(w_1w_2)$ is the predicted score when Poincar\'e embeddings are used as input.
Thus, $Score_{S}$ indicates the weighted (weight = $\alpha$) mixed prediction score from the supervised model. We measure the performance of our supervised model for each of the 25 random splits and report the mean and standard deviation of the performance metric.

\subsection{Hyperparameters of the Model} 
Apart from the hyperparameters used to train the Poincar{\'e} model, our proposed model has three hyperparameters: $k$, $m$ and $\alpha$. $k$ indicates the number of top hypernyms or hyponyms per target word to be used for training the Poincar{\'e} model. Since only considering hyponym-hypernym pairs containing target words does not lead to sufficient training samples for the Poincar{\'e} model, we add top $m$\% hyponym-hypernym pairs extracted by using Hearst pattern to the training set. Note that we consider the top hyponym-hypernym pairs on the basis of normalized frequency. $\alpha$ indicates the relative weight between Poincar{\'e} similarity and distributional similarity. We have optimized these three hyperparameters by grid search.
\section{Evaluation}

\subsection{Datasets}
\label{ds}
To evaluate our proposed models (both supervised and unsupervised) we use three gold standard datasets for English on compositionality detection and describe them in the following. 

\paragraph{Reddy (RD):} This dataset contains compositionality judgments for 90 compounds in a scale of literality from 0 (idiomatic) to 5 (compositional), obtained by averaging crowdsourced judgments on these pairs~\cite{reddy2011empirical}.  For evaluation, we use only the global compositionality score, ignoring individual word judgments. 

\paragraph{Reddy++ (RD++):} This is a recently introduced resource created for evaluation~\cite{ramisch2016naked} that extends the \text{Reddy} dataset with an additional 90 English nominal compounds, amounting to a total of 180 nominal compounds. Consistent with RD, the scores range from 0 (idiomatic) to 5 (compositional) and are annotated through Mechanical Turk and averaged over the annotators. The additional 90 entries are adjective-noun pairs, balanced with respect to compositionality.  

\paragraph{Farahmand (FD):} This dataset contains 1042 English compounds
extracted from Wikipedia with binary non-compositionality judgments by four experts~\cite{W15-0904}. In evaluations we use the sum of all the judgments  to have a single numeral compositionality score, ranging from 0 (compositional) to 4 (idiomatic). 

We optimize our method on subsets of the datasets for pairs and constituents with available Poincar{\'e} embeddings in order to measure the direct impact of our method, which comprises 79, 146 and 780 datapoints for the three sets \text{RD-R}, \text{RD++-R} and \text{FD-R}, respectively. 

We subsequently report scores on the full datasets \text{RD-F} (90), \text{RD++-F} (180) and \text{FD-F} (1042) for the sake of fair comparison to previous works. In cases where no Poincar{\'e} embeddings are available, we use the fallback strategy of only relying on the distributional model, i.e. $Score_{DS}$.

For the supervised setup, we experiment on the \text{FD} dataset (on the reduced version and the full version) since for the other two datasets, the number of instances are not enough for supervision.

\subsection{Baselines} 
We use the recent work by \citet{P16-1187} as the baseline, where authors apply several distributional semantic models and their variants by tuning hyperparameters like the dimension of vectors, the window-size during training and others. We resort to PPMI-SVD, two variants of word2vec (CBOW and SkipGram) and GloVe as our baselines. We use these models as provided, with the vector dimension size of 750 (PPMI-SVD, W2V) and 500 (GloVe)\footnote{These pre-trained DSMs were provided by~\citet{P16-1187}; on re-computation we get slightly different results than those reported in their paper.}. 

\paragraph{PPMI-SVD baseline:} For each word, its neighboring nouns and verbs in a symmetric sliding window of $w$ words in both directions, using a linear decay weighting scheme with respect to its distance $d$ to the target~\cite{levy2015improving} are extracted. The representation of a word is a vector containing the positive pointwise mutual information (PPMI) association scores between the word and its contexts. Note that, for each target word, contexts that appear less than 1000 times are discarded. The Dissect toolkit~\cite{dinu2013dissect} is then used in order to build a PPMI matrix and its dimensionality is reduced using singular value decomposition (SVD) to factorize the matrix.

\paragraph{word2vec baseline:} This DSM is prepared using the well-known word2vec  \cite{mikolov2013distributed} in both variants CBOW (W2V-CBOW) and Skip-Gram (W2V-SG), using default configurations except for the following: no hierarchical softmax; negative sampling of 25; frequent-word downsampling weight of $10^{-6}$; runs 15 training iterations; minimum word count threshold of 5.

\paragraph{GloVe baseline:} The count-based DSM of~\citet{pennington2014glove}, implementing a factorization of the co-occurrence count matrix is used for the task. The configurations are the default ones, except for the following: internal cutoff parameter $x_{max} = 75$; builds co-occurrence matrix in 15 iterations; minimum word count threshold of 5.

Other baseline models proposed by~\citet{reddy2011empirical},~\citet{salehi2014using},~\citet{salehi2015word} report results only on Reddy dataset (since the other two datasets have been introduced later) whereas ~\citet{yazdani2015learning} perform their evaluation only on the Farahmand dataset for their supervised model. In addition, this supervised approach requires an additional resource of $\sim 70k$ known noun phrases from Wikipedia for training. However, \citet{P16-1187} compare their best models with all these baseline models and show that their models outperform across all the respective datasets. Hence we execute all our evaluations by considering only the best models proposed by~\citet{P16-1187} as our baselines. 

\subsection{Evaluation Setup} 

Quantitative evaluation is usually done by comparing model outcomes against the gold standard datasets. For all the three datasets (\text{RD-R}, \text{RD++-R}, \text{FD-R}), we report Spearman’s rank correlation ($\rho$) between the scores provided by the humans and the compositionality score obtained from the models. Note that for the nominal compounds in \text{FD-R} dataset, higher human scores indicate a higher degree of idiomaticity, which is opposite to the scoring in the \text{RD-R} and \text{RD++-R} datasets. We therefore always report the absolute correlation values ($|\rho|$) for all the datasets.    

\section{Experimental Results}
\label{ER}
In this section, we report the results obtained from the baseline models and the unsupervised and supervised variants of our model. 
\subsection{Unsupervised Baseline Results} 
We compare the performance of the baseline models~\cite{P16-1187} and Poincar{\'e} embeddings as a single signal on the reduced version of the three gold standard datasets: \text{RD-R} (79 instances), \text{RD++-R} (146 instances), \text{FD-R} (780 instances) in order to closely examine the influence of Poincar{\'e} embeddings.
Table~\ref{tab:baseline_results} shows the performance for all the baselines in terms of Spearman's rank correlation $\rho$. We observe that \text{W2V-CBOW} model produces the best performance across all the three datasets and \text{W2V-SG} achieves the second-best performance. 
As noted in the table, the Poincar{\'e} embeddings on their own perform worse than all the other baselines. Further, since our final model is based on an interpolation between Poincar{\'e} embeddings and W2V-CBOW, we also attempted interpolation between other four baseline models, but the best results were always close to the better of the two models, and are not reported here. 

\begin{table}[!tbh]
\begin{center}
\small
    \begin{tabular}{|M{2.0cm}|M{1.3cm}|M{1.2cm}|M{1.6cm}|}
    \hline
    \bf{Base. Model} & \bf{RD-R} & \bf{RD++-R} & \bf{FD-R}\tabularnewline \hline

\textbf{W2V-CBOW} & \textbf{0.8045}& \textbf{0.6964}&    \textbf{0.3405}  \tabularnewline \hline 
\centering  W2V-SG & 0.8034&    0.6963&    0.3396  \tabularnewline \hline
\centering  GloVe  & 0.7604&    0.6487&    0.2620 \tabularnewline \hline
\centering  PPMI-SVD  & 0.7484&    0.6468&    0.2428 \tabularnewline \hline
\centering  Poincar{\'e}  & 0.6023&    0.4765&    0.2007 \tabularnewline \hline

    \end{tabular}

\end{center}

\caption{Baseline~\cite{P16-1187} results on the reduced version of three gold-standard datasets ordered in decreasing overall performance along with the results of using only Poincar{\'e} embedding.}
\label{tab:baseline_results}
\end{table}

\subsection{Results of Proposed Unsupervised Model}
We report the effect of tuning hyper-parameters introduced in Section \ref{model}, e.g. $k$, $m$, or $\alpha$. 

\paragraph{Fixed $k$ neighbours:}
We start by fixing $k=5$ and obtain the correlations by varying $m$ and $\alpha$. The results are presented in Table~\ref{tab:cbow+Poincare results+topm}. We experiment with values of $m$ ranging from $0$ to $10$ and report results for $m=0,1,5,10$. Note that here $m=0$ indicates the case where we use the Poincar{\'e} embeddings of the target word's top $k$ hypernyms and hyponyms only with no additional highly frequent hyponym-hypernym pairs. Values of $m>10$ degrade the quality, as too many noisy pattern extractions would be used in training.

\noindent\textit{Key observations}: For certain values of $\alpha$ we obtain considerable improvements over the baseline Spearman's correlation when introducing Poincar{\'e} embeddings. The addition of top hyponym-hypernym pairs (i.e., $m>0$) improves the performance of the model. Finally, note that for $m>0$, $\alpha=0.4$ generally produces better results across the three datasets.  

\begin{table}[!tbh]

\begin{center}\small
    \begin{tabular}{|M{0.65cm}|M{1cm}|M{1cm}|M{1.2cm}|M{1.6cm}|}
    \hline
   \emph{m}(\%)& $\alpha$ & \bf{RD-R} & \bf{RD++-R} & \bf{FD-R}\\ \hline
& \centering  0.2 & 0.8160    & 0.7102    & 0.3536 \\ \cline{2-5} 
0 & \centering  0.4  &  0.8117    &  0.7012 &    0.3532\\ \cline{2-5}
& \centering  0.6  & 0.7844    & 0.6581 &    0.3278\\ \hline
   
&  0.2 & 0.8274    & 0.7155    & 0.3482 \\ \cline{2-5} 
1 & \bf 0.4  & \bf 0.8391    & \bf 0.7165 &    \bf0.3373 \\ \cline{2-5}
& 0.6  & 0.8136    & 0.6817 &    0.3036\\ \hline

& \centering  0.2 & 0.8362    & 0.7268 &    0.3501 \\ \cline{2-5} 
5 &\bf 0.4  & \bf 0.8578    & \bf 0.7389 &    \bf 0.3432\\ \cline{2-5}
&  0.6  & 0.8467    & 0.7279 &    0.3126\\ \hline

& \centering  0.2 & 0.8346 &    0.7250 &    0.3513 \\ \cline{2-5} 
10 & \centering  \bf 0.4  & \bf 0.8421    & \bf 0.7461 &    \bf 0.3469\\ \cline{2-5}
& \centering  0.6  & 0.8299    & 0.7372 &    0.3204\\ \hline

    \end{tabular}
\end{center}
\caption{Effect of the introduction of the Poincar{\'e} embeddings for varying values of $m$ and $\alpha$. Here W2V-CBOW is used as distributional model.}
\label{tab:cbow+Poincare results+topm}
\end{table}

\begin{table}[!tbh]
\begin{center}\small
    \begin{tabular}{|M{1cm}|M{1.3cm}|M{1.2cm}|M{1.8cm}|}
    \hline
\multicolumn{4}{|c|}{MODEL-DP with W2V-CBOW}\\ \hline
   $\alpha$ & \bf{RD-R} & \bf{RD++-R} & \bf{FD-R}\\ \hline
\centering  0.2 & 0.8265    & 0.7177 &    0.3594 \\ \hline 
\centering  \bf 0.4  & \bf 0.8324    & \bf 0.7321    & \bf 0.3646\\ \hline
\centering  0.6  & 0.8082    & 0.7077    & 0.3450\\ \hline 
\hline
\multicolumn{4}{|c|}{MODEL-DP with W2V-SG}\\ \hline
   $\alpha$ & \bf{RD-R} & \bf{RD++-R} & \bf{FD-R}\\ \hline
\centering  0.2 & 0.8244    & 0.7215 & 0.3603 \\ \hline 
\centering  \bf 0.4  & \bf 0.8330    & \bf 0.7337    & \bf 0.3673\\ \hline
\centering  0.6  & 0.8152    & 0.7101    & 0.3461\\ \hline

    \end{tabular}
\end{center}
\caption{Performance of MODEL-DP using W2V-CBOW as well as W2V-SG as distributional models: Effect of removal of top 1\% hypernym-hyponym pairs from the top 10\% pairs ($k=5$).}
\label{tab:cbow_results}
\end{table}

\paragraph{Effect of the top $m$ pairs:} Since the extraction of the hypernyms from the corpus is completely unsupervised and based on handcrafted lexical-syntactic patterns, we investigate whether the most frequent hyponym-hypernym pairs are affecting the quality of Poincar{\'e} embeddings, having noted many erroneous extractions for very frequent pairs. We fix the value of $m=10$, but drop the most frequent 1\% hyponym-hypernym pairs and retrain the Poincar{\'e} model with the rest of the pairs. We call this variant \text{MODEL-DP}. The upper half of Table~\ref{tab:cbow_results} shows the performance of this model while using W2V-CBOW as the distributional models ($k=5$, which was the optimal $k$ also in this setting). We compare the result of \text{MODEL-DP} for $\alpha=0.4$  with Table~\ref{tab:cbow+Poincare results+topm}, row corresponding to $m=10\%,\alpha=0.4$.

\begin{table}[!tbh]
\begin{center}
\small
    \begin{tabular}{|M{0.5cm}|M{0.8cm}|M{1.3cm}|M{1.5cm}|M{1.3cm}|}
    \hline
   \emph{k}& $\alpha$ & \bf{RD-R} & \bf{RD++-R} & \bf{FD-R}\\ \hline
& \centering  0.2 & 0.8269        & 0.7228 &     0.3563 \\ \cline{2-5} 
3 & 0.4  & 0.8275 & 0.7382 & 0.3557\\  \cline{2-5}
\cline{2-5}
& \centering  0.6  & 0.8089 & 0.7188 & 0.3278    \\ \hline \hline
& \centering  0.2 & 0.8265    & 0.7177 &    0.3594 \\ \cline{2-5} 
\bf 5 & \centering  \bf 0.4  & \bf 0.8324    & \bf 0.7321    & \bf 0.3646\\ \cline{2-5}
& \centering  0.6  & 0.8082    & 0.7077    & 0.3450\\ \hline \hline

& \centering  0.2 & 0.8123    & 0.7103 & 0.3534 \\ \cline{2-5} 
10 & \centering   0.4  & 0.8168    &     0.7248    &  0.3589\\ \cline{2-5}
& \centering  0.6  & 0.7700 & 0.6957 & 0.3484\\ \hline 
    \end{tabular}
    \end{center}

\caption{Results obtained for MODEL-DP ($m=10$, top 1\% hypernym-hyponym pairs removed) by varying the values of $k$.}
\label{tab:cbow+Poincare results+topk}

\end{table}

\noindent\textit{Key observations}: We mainly observe that discarding the most frequent 1\% hyponym-hypernym pairs improves the results for the largest dataset  \text{FD-R} considerably while making the results from the other two datasets a little worse. We also produce results on MODEL-DP by varying the value of $k$. We try with $k=3,5,10$, the results of which is presented in Table~\ref{tab:cbow+Poincare results+topk}. Clearly, $k=5$ gives the best performance.

If we consider very few hypernyms per target word, it results in lack of sufficient information for the Poincar{\'e} model, while training with too many hypernyms per target word dilutes the useful hierarchy information because it adds noise.

\paragraph{Other DSM models:} We use W2V-CBOW as the DSM for MODEL-DP. %In the MODEL-DP the DSM used is W2V-CBOW. 
Keeping all the other parameters of MODEL-DP the same (i.e., $m=10$, $k=5$, $\alpha=0.4$) we replace the DSM by the W2V-SG vectors, which was performing the second best among the baselines. We are interested in observing whether the Poincar\'e embeddings also benefit other DSM models as well.

\noindent\textit{Key observations}: The performance of this variant of our model is presented in the lower half of Table~\ref{tab:cbow_results}. We indeed observe the same effect of the Poincar\'e embeddings improving the overall performance by 3-4\% on all datasets. 

\paragraph{Other hyperparameters:} In a series of experiments that we do not report in detail for brevity, we could make the following observations: For our task, the vector dimensionality of Poincar{\'e} embeddings of $d = 50$ shows better results than higher or lower values, as tested with $d \in \{20,100\}$. Similarly, we tried with several vector dimensions of DSMs with $d \in {50,100,300}$ but 750 gives the best performance for the best models reported by~\citet{P16-1187} and our model in the unsupervised setup.  We further tried varying the relative weight of single word vectors for the sum in Equation~\ref{eq1}, which did not have positive effects.

\begin{table}[!tbh]
\begin{center}
\small
    \begin{tabular}{|M{2cm}|M{1cm}|M{1.4cm}|M{1.6cm}|}
    \hline
    \multicolumn{4}{|c|}{Performance for reduced dataset }\\ \hline
    Model & \bf{RD-R} & \bf{RD++-R} & \bf{FD-R}\\ \hline
\centering W2V-CBOW & 0.8045 & 0.6964 & 0.3405 \\ \hline 
\centering MODEL-DP& \bf 0.8324 & \bf 0.7321 & \bf 0.3646
   \\ \hline \hline
\multicolumn{4}{|c|}{Performance for full dataset }\\ \hline 
    Model & \bf{RD-F} & \bf{RD++-F} & \bf{FD-F}\\ \hline

\centering W2V-CBOW & 0.7867 & 0.7022 & 0.2688 \\ \hline 
\centering MODEL-DP& \bf 0.8095 & \bf 0.7302 & \bf 0.2958
   \\ \hline 

    \end{tabular}
\end{center}

\caption{Performance of our model (MODEL-DP) and most competitive baseline (W2V-CBOW) for both the reduced datasets and the whole datasets (using the fallback strategy).}
\label{tab:cbow_relaxed_results}

\end{table}

\paragraph{ Fallback strategy to encompass the whole dataset:} In all the above experiments we consider the reduced version of the three gold-standard datasets due to lack of the Poincar\'e embeddings for certain target words. We suggest a fallback strategy to incorporate the target words that do not have Poincar\'e embeddings. In cases where the Poincar\'e embeddings are not present, we fall back to the distributional similarity score. In cases, where the Poincar\'e embeddings are available we use the combined score as discussed in Section~\ref{model}. Note that, the distributions of distributional similarity scores and proposed combined scores are significantly different (according to the $z$-test~\cite{fisher1932statistical}). Therefore while falling back to the distributional similarity scores we scale up the scores by the proportion of normalized means of the two distributions.

\noindent\textit{Key observations}: The results for this fall back strategy is noted in the lower half of Table~\ref{tab:cbow_relaxed_results}. We observe that for all three datasets we perform significantly better than the baselines. To be consistent with the literature, we compare our performance even with the supervised model proposed by~\citet{yazdani2015learning} \iffalse both in terms of Spearman's rank correlation ($\rho$) and best F1 score~\cite{yazdani2015learning}\fi for the \text{FD-F} dataset. For this dataset, the supervised model proposed by the authors produces a Spearman's rank correlation ($\rho$) of 0.41 whereas the unsupervised MODEL-DP produces 0.29. However, our supervised approach, as we shall see later, beats this number reported by~\citet{yazdani2015learning} by a considerable margin. 

\paragraph{Significance test:} From the extensive evaluation of our model by tuning several hyper-parameters, we obtain MODEL-DP (Table~\ref{tab:cbow_results}), which gives the best performance for all the three datasets outperforming the baselines (Table~\ref{tab:baseline_results}). We perform Wilcoxon's sign-rank test \cite{rey2011} for all the three datasets separately. We obtain $p < 0.05$ while comparing MODEL-DP and the best baseline model (W2V-CBOW) indicating that the difference between their compositionality predictions is statistically significant.

\paragraph{Error analysis:} We investigate the erroneous cases for which the annotators give a high compositionality score while our model produces a very low compositional score, e.g. `area director', `discussion page', and `emergency transportation'. We observe that the number of hypernyms extracted for these target noun phrases is very low (1 or 2), which leads to a less informative hierarchical representation in the Poincar\'e model; this is either caused by a low frequency of terms overall, or by a low occurrence in hypernym pattern contexts. 
We also analyzed the non-compositional cases for which the annotators give a low compositionality score but our model produces a high score, e.g. `hard disk', `hard drive' and `soft drink'. In these cases even though they are non-compositional, the hypernyms of the noun phrases match with the hypernyms of the head constituent words. For example, `hard disk' and `disk' have the same hypernym `storage device'; similarly `soft drink' and `drink' have `product'; `hard drive' and `drive' have `device'. Thus, these non-compositional cases are different from entirely opaque expressions like `couch potato', `hot dog' where none of the hypernyms of the noun phrases match with the hypernyms of any of the constituent words. Categorizing the non-compositional words based on the above observation and dealing with such cases is left for future work. 

\paragraph{Training using lexical resources:}
We further investigated the use of hyponym-hypernym pairs extracted from lexical resources like WordNet~\citep{wordnet:1993} or ConceptNet~\citep{AAAI1714972} for training the Poincar{\'e} model. Even though the quality of the hyponym-hypernym pairs from lexical resources is better compared to the pairs extracted using Hearst patterns, the coverage of target words is very low. Therefore, for a fair comparison, we prepare a reduced version of the three gold standard datasets (RD-RL, RD++-RL, FD-RL), where all the target words are present in lexical resources as well as hyponym-hypernym pairs extracted using Hearst patterns. RD-RL, RD++-RL, and FD-RL contain 74, 131, 380 target words, respectively. MODEL-DP-L uses the same compositionality score metric as MODEL-DP but in the case of MODEL-DP-L, the Poincar{\'e} embedding is learned using the hyponym-hypernym pairs extracted only from WordNet and ConceptNet combined. The results are presented in Table~\ref{tab:lex_results}. We see that even though MODEL-DP-L performs better than the baselines for two of the datasets, MODEL-DP gives the best result. We attribute this to the relative sparsity of lexical resources, which are seemingly not sufficient for training reliable Poincar{\'e} embeddings.     

\begin{table}

\begin{center}
\small
    \begin{tabular}{|M{2cm}|M{1.28cm}|M{1.4cm}|M{1.28cm}|}
    \hline
    Model & \bf{RD-RL} & \bf{RD++-RL} & \bf{FD-RL}\\ \hline

\centering W2V-CBOW  & 0.8111 & 0.7256 & 0.4198  \\ \hline 
\centering MODEL-DP-L& 0.8223 & 0.7451 & 0.4179 \\ \hline
\centering MODEL-DP& \bf 0.8288 & \bf 0.7592 & \bf 0.4790   \\ \hline 

    \end{tabular}
\end{center}
\caption{Comparisons of the results produced by MODEL-DP-L from lexical resources vs. MODEL-DP along with the baselines for the reduced dataset.}
\label{tab:lex_results}
\end{table}

\begin{table}[!tbh]

\begin{center}

\resizebox{0.98\columnwidth}{!}{
    \begin{tabular}{|M{1.9cm}|M{1.3 cm}|M{0.9cm}|M{1.3 cm}|M{0.9cm}|}
    \hline
   \multicolumn{5}{|c|}{\bf{FD-R}} \\\cline{1-5}
     &\multicolumn{2}{|c|} {Kernel Regression} & \multicolumn{2}{|c|}{PLS Regression} \\ \hline

& $\mu(|\rho|)$ & $\sigma(|\rho|)$ & $\mu(|\rho|)$ & $\sigma(|\rho|)$ \\\hline
CBOW-S (750) & {0.4017} & 0.0599 & 0.3972 & 0.0590 \\ \hline 

$\alpha$ &\multicolumn{4}{|c|}{MODEL-DP-S} \\ \hline  
0.2 & 0.4294 & 0.0591 & 0.4078 & 0.0566 \\ \hline 

\bf 0.4 & \bf 0.4347 & \bf 0.0563 & \bf 0.4096 & \bf 0.0525  \\ \hline 

0.6 & 0.4221 & 0.0540 & 0.3959& 0.0497  \\ \hline \hline 
CBOW-S (50) & {0.4339} & 0.0570 & 0.4227 & 0.0584 \\ \hline 
$\alpha$ & \multicolumn{4}{|c|}{MODEL-DP-S, CBOW vectors of dim. 50} \\ \hline
0.2 & 0.4487 & 0.0547 & 0.4361 & 0.0561 \\ \hline 
\bf 0.4 & \bf 0.4520 & \bf 0.0528 & \bf 0.4372 & \bf 0.0518  \\ \hline 
0.6 & 0.4410 & 0.0510 & 0.4196 & 0.0491  \\ \hline \hline

\multicolumn{5}{|c|}{\bf{FD-F}} \\\cline{1-5}
    &\multicolumn{2}{|c|} {Kernel Regression} & \multicolumn{2}{|c|}{PLS Regression} \\ \hline

& $\mu(|\rho|)$ & $\sigma(|\rho|)$ & $\mu(|\rho|)$ & $\sigma(|\rho|)$ \\\hline
CBOW-S (750) & 0.3822 &0.0471 & 0.3910 & 0.0434 \\ \hline 
$\alpha$ & \multicolumn{4}{|c|}{MODEL-DP-S} \\ \hline
0.2 & 0.4030 & 0.0446 & 0.3984 & 0.0450 \\ \hline 
\bf 0.4 & \bf 0.4083 & \bf 0.0425 & \bf 0.3941 & \bf 0.0459  \\ \hline 
0.6 & 0.3986 & 0.0418& 0.3747& 0.0471  \\ \hline \hline
CBOW-S (50) & {0.4212}&0.0502&0.4201&0.0470 \\ \hline 
$\alpha$ & \multicolumn{4}{|c|}{MODEL-DP-S, CBOW vectors of dim. 50} \\ \hline
0.2 & 0.4329 & 0.0500 & 0.4270 & 0.0467 \\ \hline 
\bf 0.4 & \bf 0.4340 & \bf 0.0488& \bf 0.4211& \bf 0.0469  \\ \hline 
0.6 & 0.4213 & 0.0478& 0.3943& 0.0499  \\ \hline

    \end{tabular} 
    }
\end{center}
\caption{Mean ($\mu$) and Standard Deviation ($\sigma$) of Spearman's rank correlation ($\rho$) of the supervised approach for \text{FD-R} and \text{FD-F} datasets over 25 random splits. We compare best baseline model (CBOW - 750 and 50 dimension) and our model (MODEL-DP-S) using both 750 and 50 dimension of CBOW vectors.}
\label{tab:supervised_results}
\end{table}
\subsection{Results of Proposed Supervised Model}
For the supervised setup we present our results on the reduced \text{FD-R} dataset (780 instances) and the full Farhamand \text{FD-F} dataset (1042 instances). We do not use the other two datasets for the supervised setup since the number of instances in both these datasets are too small to produce a reasonable training-test split required for supervision.

As discussed in Section~\ref{sp}, we use various regression models; 75\% of the dataset is used for training and the remaining 25\% is used for testing; we experiment on 25 such random splits and report mean and standard deviation of Spearman's rank correlation ($\rho$). Among all the regression models (respective to the best choice of the hyperparameters),  Kernel Ridge regression gives the best performance while PLS regression is the second best for both the \text{FD-R} and \text{FD-F} dataset. We compare the performance of the best baseline supervised model (CBOW-S) where only $Score_{DS}$ from Equation~\ref{se} is used as the predicted score with our proposed supervised model (MODEL-DPS) where $Score_{S}$ from Equation~\ref{se} is used as the predicted score. The performance of these two best regression models for the baseline and our model (for $\alpha=0.4$)\footnote{$\alpha=0.4$ produces the best results per grid search.} are noted in Table~\ref{tab:supervised_results}.  In the same table, we also report the results of the evaluation on \text{FD-F} dataset using a fallback strategy for the supervised setup: here, we use a 50-dimensional zero vector of the target word or compound for which the Poincar{\'e} embedding is absent. 
We observe that for both the datasets (reduced and full) our approach outperforms the baseline results by a large margin. 
As discussed earlier, the CBOW vectors used for experiments consist of 750 dimensions. Since the number of data points in the training set is small, we also experiment with CBOW vector dimension of 50 (MODEL-DPS-50) in the supervised setup to avoid overfitting due to a large number of parameters. The results presented in Table~\ref{tab:supervised_results} show that with the reduced number of dimensions, our model yields even better results and outperforms the correlations 0.41 and 0.34 reported respectively  in~\cite{yazdani2015learning} and~\cite{P16-1187}.    

\section{Conclusion}
In this paper, we present a novel straightforward method for estimating degrees of compositionality in noun phrases. The method is mixing hypernymy and distributional information of the noun phrases and their constituent words. To encode hypernymy information, we use Poincar{\'e} embeddings, which -- to the best of our knowledge -- are used for the first time to accomplish the task of compositionality prediction. While these hyperbolic embeddings trained on hypernym pattern extractions are not a good signal on their own for this task, we observe that mixing distributional and hypernymy information via Euclidean and hyperbolic embeddings helps to substantially and significantly improve the performance of compositionality prediction,  outperforming previous state-of-the-art models. 
Our pretrained embeddings and the source codes are publicly available.\footnote{\url{https://github.com/uhh-lt/poincare}}

Two directions for future work are (i) to extend our approach to other languages by using multilingual resources or translation data; and (ii) to  explore various compositionality functions to combine the words' representation on the basis of their grammatical function within a phrase.

\subsubsection*{Acknowledgments}

We acknowledge the support of the DFG under the ``JOIN-T'' (BI 1544/4) and ``ACQuA'' (BI 1544/7) projects, Humboldt Foundation for providing scholarship as well as the DAAD and the Indian Department of Science and Technology via a DAAD-DST PPP grant. 

%\bibliographystyle{acl_natbib}
%\bibliography{acl2019}

%\if{0}
\Urlmuskip=0mu plus 1mu\relax

%\fi

\end{document}